\pdfoutput=1

\documentclass[11pt]{article}

\usepackage[table,xcdraw]{xcolor}
\usepackage[]{emnlp2021}

\usepackage{times}
\usepackage{latexsym}

\usepackage[T1]{fontenc}

\usepackage[utf8]{inputenc}

\usepackage{microtype}

\usepackage{graphicx}
\usepackage{bm}
\usepackage{multirow}
\usepackage{cjhebrew}

\title{Neural Token Segmentation for High Token-Internal Complexity}

\author{Idan Brusilovsky \\
  The Open University \\
  \texttt{brusli1@gmail.com} \\\And
  Reut Tsarfaty \\
  Bar Ilan University \\
  \texttt{reut.tsarfaty@biu.ac.il} \\}

\begin{document}
\maketitle
\begin{abstract}

Tokenizing raw texts into word units is an essential pre-processing step for critical tasks in the NLP pipeline such as tagging, parsing, named entity recognition, and more.
For most languages, this tokenization step straightforward.  However, for languages with high token-internal complexity, further token-to-word segmentation is required. Previous canonical segmentation studies were based on character-level  frameworks, with no contextualised representation involved.  Contextualized vectors \`{a} la BERT show remarkable results in many applications, but were not shown to improve performance on linguistic segmentation  per se. Here we propose a novel neural segmentation model which combines the best of both worlds, contextualised token representation and char-level decoding, which is particularly effective for languages with high token-internal complexity and extreme morphological ambiguity.  Our model shows substantial  improvements in segmentation accuracy on Hebrew and Arabic compared to the  state-of-the-art, and leads to further improvements on downstream tasks such as Part-of-Speech Tagging, Dependency Parsing and Named-Entity Recognition, over existing pipelines. 
When comparing our segmentation-first pipeline with joint segmentation and labeling in the same settings, we show that, contrary to pre-neural studies, the pipeline performance is superior.
\end{abstract}

\section{Introduction}
Tokenization refers to the process of splitting raw, space-delimited, tokens into distinct meaning-bearing units. %
A case in point is {\em universal dependencies} (UD) \cite{nivre-etal-2020-universal} which adopt a two-level scheme,\footnote{See the UD guidelines: \url{universaldependencies.org/u/overview/tokenization.html}.} where tokens are first split into basic word-units, and only then further analyses (POS, dependencies) of these units is provided.

Specifically, in UD and  many other NLP tasks the assumption is that a single word-unit needs to correspond to a single Part-of-Speech tag. In most languages, the process of extracting  such word-units from space-delimited tokens is  straightforward.
In English, for instance, this involves
splitting `{\em isn't}' into `{\em is}' + `{\em n't}' or `{\em John's}' into `{\em John}' + `{\em 's}', which   is a deterministic and unambiguous  process for the most part. 
But for languages with high token-internal complexity and ambiguity, such as Hebrew and Arabic, this is not quite so. 
The rich  orthographic and morpho-phonological processes that form  tokens in these languages, as well as the lack of
vocalization (a.k.a.\ diacritics, {\em nikkud}) in their texts, leads to extreme token-level ambiguity that  poses particular challenges to segmentation.
Consider, for instance, the Hebrew token \cjRL{b.slm}. It could map to: \cjRL{b--.sl--/sl--hm} (literally: in-shadow-of-them, meaning: in their shadow), \cjRL{b.sl--/sl--hm} (literally: onion-of-them, meaning: their onion), \cjRL{b--h--.slm} (literally: in-the-image, meaning: in the image) and  more. Out of context, all of these analyses are equally likely.  The correct segmentation    becomes available only  in  the greater context of the global interpretation of the sentence.

Because segmentation\footnote{Following \cite{more-etal-2019-joint, article, nivre-etal-2020-universal, shao-etal-2017-character}, we use the term {\em segmentation} for the task of extracting {\em word}-units from tokens. This task is different from {canonical segmentation} in \newcite{kann-etal-2016-neural}, where canonical segments refer to {\em morphemes}.} 
for these languages is {\em critical}, many previous efforts on segmentation were language-specific \cite{monroe-etal-2014-word,article, zalmout2017don, samih-etal-2017-neural, 8620203, sajjad-etal-2017-challenging, tawfik-etal-2019-morphology}. Other efforts such as UDPipe \cite{udpipe:2017}, Stanza \cite{qi2020stanza}, \newcite{shao-etal-2017-character} or  Morfessor \cite{creutz-lagus-2002-unsupervised}, aimed at {\em universal} segmentation models  which are language agnostic. %
However, on widely accepted cross-lingual benchmarks  as UD, their performance on  languages with complex  ambiguous tokens (see \textsection\ref{sec:res}) lags behind.
On top of that, recent prominent works on canonical segmentation of morphologically-complex languages \cite{kann-etal-2016-neural, qi2020stanza, DBLP:journals/corr/abs-1807-02974} utilized character-level sequence to sequence frameworks, yet lacked the {\em critical} disambiguiating context of the tokens, as required by cases of extreme token-internal ambiguity.

In this paper, we propose to bridge this critical gap by devising a {\em  char-token segmentation} (CATS)  model\footnote{All of our code and models will be made publicly available upon acceptance at {\em anonymous.com}.} where a character-based encoder-decoder network with attention is designed to take advantage of both the token's surface form via character representations, and the full token's   {\em  contextualized} embedding. 
The model is trained end-to-end, is completely language-agnostic, and does not require any external symbolic resources (contrary to  \newcite{zalmout2017don,seker-tsarfaty-2020-pointer,seeker-cetinoglu-2015-graph}, and others). 

We applied our model to a set of languages of high token-internal complexity from the Universal Dependencies 2.5 (UD)  project \cite{nivre-etal-2020-universal}, outperforming state-of-the-art segmentation results on Hebrew, Arabic and Turkish. %
We confirm the utility of our segmentation  on three downstream tasks: Part-Of-Speech (POS) tagging, dependency parsing and Named-Entity Recognition (NER), with substantial improvements over existing pipelines.
Furthermore, we observed that, contrary to pre-neural studies \cite{cohen2007joint, adler2006unsupervised,  seker-tsarfaty-2020-pointer,10.1162/tacl_a_00404} we see no particular advantage for joint modeling of segmentation and labeling over the pipeline in these settings.

\section{Models}

Let us begin by defining the {\em contextualized segmentation}
 task we are interested in. Formally, let  $\mathcal{V}$ be the token vocabulary of a given language and $\mathcal{V}_w$ be the word vocabulary. 
We aim to induce a function \(f(v,c)=W\) that finds
for every token $t \in \mathcal{V}$ in a given context $c \in C$ the list of  words $W = (w_1,...,w_n\mid w_i \in  \mathcal{V}_w)$ composing this token.
In practice, this means that the model's input consists of the character-sequence of the current surface token, and a %
{\em  contextualized}  representation  of this token within the sentence. The  output is a  character sequence  where the \textit{space} symbol indicates separation between words.

\begin{figure}[t]
    \centering
    \includegraphics[width=0.45\textwidth]{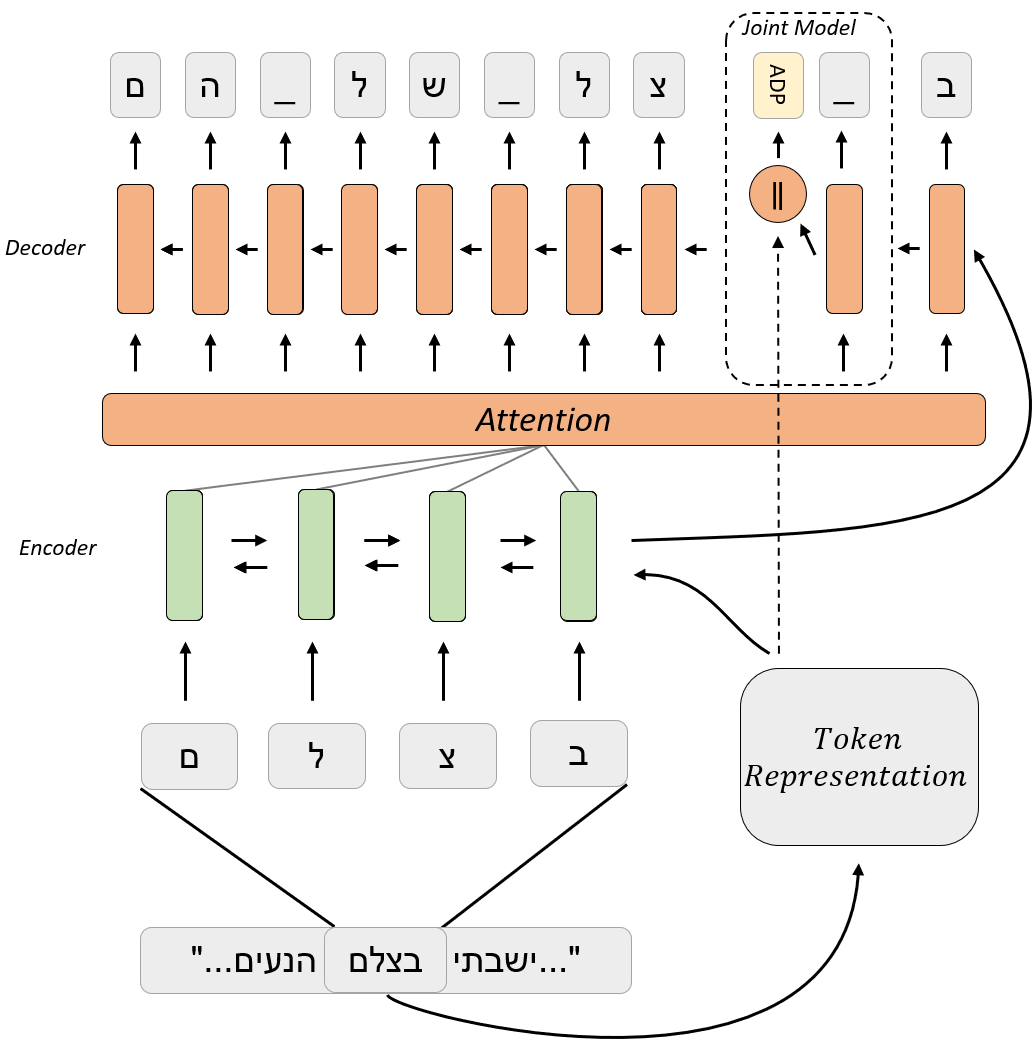}
    \caption{The CATS architecture. The Input consists of char representations augmented with token representation. Green indicates the BiLSTM encoder. Orange is the LSTM-with-attention decoder. The extension to a joint model (J-CATS) is illustrated in the dashed box.}
    \label{fig:architecture}
\end{figure}

Our neural segmentation model (Figure \ref{fig:architecture}) is a char-to-char attention-based encoder-decoder network architecture \cite{vaswani2017attention} reminsicent of the work of  \newcite{kann-etal-2016-neural} on canonical segmentation. In contrast with \newcite{kann-etal-2016-neural, udpipe:2017} and others, the char-based representation is extended with the full token's embedding.
The encoder is  a single-layer character-based BiLSTM fed also  with the token representation, which, in the case of a contextualized representation, takes into account the entire sentence. Our model is agnostic as to the choice of  token vector embeddings, and in Sec.\ \textsection\ref{contexts} we plug-in and compare different alternatives.
Our decoder is a char-based LSTM  with  underlying attention. %
The LSTM output enters a linear classification head with softmax activation, predicting  the next character: a letter, a \textit{space} symbol, or end-of-token (EOT). We train the network using Cross-Entropy loss.

We further extend the architecture to perform joint segmentation and labeling. We retain the same  architecture
and add an extra classification head on top of the decoder's LSTM states, concatenated with the input's token vector (Figure \ref{fig:architecture}), emitting the label predictions on the condition that a \textit{space} was generated. The network's loss is given by
\(L = \lambda\cdot L_{seg} + (1-\lambda) \cdot L_{tag}\),
where $L_{seg}$ and $L_{tag}$ are the character prediction and tagging Cross-Entropy losses, and $\lambda$ is a hyper-parameter between $(0,1)$ tuned on the \textit{dev} set.

\section{Experiments}
\paragraph{Goal} \label{morphplogical_segmentation}
We set out to empirically evaluate the proposed model on  languages of high token-internal complexity from the UD 2.5 datasets \cite{nivre-etal-2020-universal}.
 We use the standard \textit{train/dev/test} split of the UD corpora and report   results on the \textit{test} set of each language. 

\paragraph{Segmentation Settings} \label{contexts}

In our experiments we use  two different types of contextualized token embeddings 
     (i)  \textbf{RNN-based:}
      applying a single-layer BiLSTM, and using the BiLSTM outputs as contextualized representations.
      The token embeddings are frozen, and the BiLSTM is trained as part of the model.
      (ii)  \textbf{Transformer-based:}
      We use multi-lingual BERT \cite{devlin2018bert} hosted by \newcite{Wolf2019HuggingFacesTS} to extract contextualized embeddings for all tokens. In order to obtain the token's contextualized vector, we  average the vectors of all  word pieces of that  token. 

All Models were trained with an Adam Optimizer with learning rate of $1e^{-3}$, and batch size of 128. Each dataset was trained for a different amount of epochs, depending on its size (20 to 40).

\paragraph{Pipeline Settings}
We evaluate our segmentation on three downstream tasks: POS tagging, dependency parsing and NER. 
We experimented with four types of segmentations in our pipeline setup:
   (i) {\em Oracle:} The gold segmentation given by UD.
     (ii) {\em UDPipe:} The predicted segmentation by UDPipe.
     (iii) {\em Stanza:} A multi-word segmentation tool by Stanza \cite{qi2020stanza}.
    (iv) {\em CATS:}  our proposed  segmentation model. For all languages, we rely on mBERT  embeddings \cite{devlin2018bert}.

Next, for the POS and NER tasks, the segmented text is followed by a BERT-based token classification model \cite{Wolf2019HuggingFacesTS}. The classification model consists of a single classification head on top of a BERT model, fine-tuned on  \textit{train}  for 3 epochs with batch size of 8. For dependency parsing we used the parsing capability of Stanza.

For the POS tagging and depedency parsing tasks, the same 4 datasets of UD 2.5 are used, using the UPOS column as gold for the POS task. For  NER, a gold pre-segmented NER dataset  is required for training and evaluation, so we resorted to a Hebrew dataset  which consists  \textit{BIOSE} NE labels on top of the  segmented raw tokens  \cite{bareket2020neural}.

\paragraph{Joint Settings}
We  evaluated our joint model on both POS and NER tasks, with loss weight of $\lambda = 0.2$. Since NER is a task that requires more {\em semantic} information (as opposed to syntactic POS tags) we experimented with a model variant (nicknamed JS-CATS) concatenating the BERT {\em sentence} embedding (i.e., the <CLS> token representation) on top of the token vector.

\paragraph{Baselines}
We use three  kinds of baselines:\\
\textbf{(i) No-Contextualization Baslines:}  To examine the contribution of the {\em pre-trained} token embeddings, we test our model with  {\em non}-contextualized token embeddings (initialized either by Zeros or using FastText (FT)) trained with the main task (essentially falling back on standard canonical segmentation architecture as in \cite{kann-etal-2016-neural}).
\\\textbf{(ii) No-Char Baselines:} We   test the multi-lingual BERT \cite{devlin2018bert} segmentation capabilities by, first, testing its internal word-piecess tokenizer, and also using simple LSTM character decoder with only the BERT vector as its input encoding,  trained with the main task. 
In addition we compare our models to  current state-of-the-art: 
\textbf{(iii) Language-Agnostic SOTA:} We compare our models to the language-agnostic segmentation models of UDPipe \cite{udpipe:2017} and Stanza \cite{qi2020stanza}.\footnote{The model of \newcite{DBLP:journals/corr/abs-1807-02974} fit this criteria, however, the code  is obsolete and not reproducible.}\footnote{For completeness, in the supplementary material we added \textbf{(iv) Language-specific SOTA:}     comparing our model to the state-of-the-art language-specific model of Hebrew segmentation,  YAP \cite{more-etal-2019-joint}, which is known to be state-of-the-art on the  Hebrew section of the SPMRL shared task \cite{seddah-etal-2014-introducing} and is  the de-facto standard for segmentation work on Hebrew, in both academia and the industry.}

\paragraph{Evaluation} \label{seg-eval}
To evaluate segmentation, we adopt the  precision, recall and F1-score, defined by \newcite{DBLP:journals/corr/abs-1807-02974}.\footnote{Using the evaluation code of \newcite{DBLP:journals/corr/abs-1807-02974} .} %
We  compute the metrics on the set of predicted surface segments, compared to the gold surface segments. %
We evaluate POS tagging  by redefining the segments to include their POS labels and calculate F1-scores as usual. For NER we use  the standard method of calculating F1-scores over entity spans, respecting both their surface form and label,  as explained in \newcite{bareket2020neural}. For dependency parsing, standard UAS/LAS scores do not fit the task since predicted segmentation may differ from the gold sequence, leading to indices mismatch. We thus use the {\em aligned multi-set} F1-score on Form-Head-Relation triplets.

\section{Results}
\label{sec:res}
Table \ref{seg_results} presents the segmentation F1-scores on all models for the UD languages we experiment with. 
First, we observe that BERT alone cannot cope with the complex segmnetation of multi-word tokens -- neither using its internal tokenizer, nor via its token-based vector embeddings.
Further, our contextualized models show substantial improvements in segmentation scores on Hebrew, Arabic and Turkish, compared to all baselines and previous SOTA.  Contextualized token embeddings exceed the performance of non-contextualized ones,  highlighting the importance of context for diambiguation. All in all, both the token's form and the context contribute to segmentation accuracy.

\begin{table}[t]
\scalebox{0.83}{
\centering
\begin{tabular}{l|c|c|c|}
\cline{2-4}
                                  & \begin{tabular}[c]{@{}c@{}}Hebrew\\ HTB\end{tabular} & \begin{tabular}[c]{@{}c@{}}Arabic\\ PADT\end{tabular} & \begin{tabular}[c]{@{}c@{}}Turkish\\ IMST\end{tabular} \\ \hline\hline
\multicolumn{1}{|l|}{BERT Tokens} & 38.8                                                 & 36.13                                                 & 28.98                                                  \\ \hline
\multicolumn{1}{|l|}{BERT decode} & 65.88±0.76                                           & 74.46±0.1                                             & 55.52±0.83                                             \\ \hline
\multicolumn{1}{|l|}{UDPipe}      & 85.22                                                & 94.58                                                 & 98.31                                                  \\ \hline
\multicolumn{1}{|l|}{Stanza}      & 93.19                                                & 97.88                                                 & 98.07                                                  \\ \hline\hline
\multicolumn{1}{|l|}{CATS: Zeros} & 94.03±0.15                                           & 98.3±0.13                                             & 97.6±0.56                                              \\ \hline
\multicolumn{1}{|l|}{CATS: FT}    & 95.76±0.18                                           & 98.5±0.05                                             & 97.72±0.2                                              \\ \hline
\multicolumn{1}{|l|}{CATS: RNN}   & 95.59±0.2                                            & \textbf{98.69±0.15}                                   & 97.9±0.29                                              \\ \hline
\multicolumn{1}{|l|}{CAST: BERT}  & \textbf{95.84±0.29}                                  & 98.57±0.15                                            & \textbf{98.43±0.19}                                    \\ \hline
\end{tabular}
}
\caption{Segmentation F1-scores for All Models: Mean and Standard Deviation over 5 runs.}
\label{seg_results}
\end{table}

\begin{table}[t]
\centering
\scalebox{0.65}{
\begin{tabular}{cc|c|c|c|c|c|}
\cline{3-7}
\multicolumn{1}{l}{}                                   & \multicolumn{1}{l|}{} & Oracle                        & UDPipe                                 & Stanza                        & CATS                                   &                                                                    \\ \cline{3-6}
\multicolumn{1}{l}{}                                   &                       & \multicolumn{4}{c|}{BERT Sequence Labeling}                                                                                                     & \multirow{-2}{*}{\begin{tabular}[c]{@{}c@{}}J\\ CATS\end{tabular}} \\ \cline{3-7} 
\multicolumn{1}{l}{}                                   &                       & \multicolumn{4}{c|}{Stanza Parsing}                                                                                                             & \multicolumn{1}{l|}{}                                              \\ \hline
\multicolumn{1}{|c|}{}                                 & SEG                   & 100.00                        & 85.22                                  & 93.19                         & \textbf{95.84}                         & 95.08                                                              \\ \cline{2-7} 
\multicolumn{1}{|c|}{}                                 & POS                   & \cellcolor[HTML]{EFEFEF}96.49 & \cellcolor[HTML]{EFEFEF}83.13          & \cellcolor[HTML]{EFEFEF}89.86 & \cellcolor[HTML]{EFEFEF}\textbf{92.73} & \cellcolor[HTML]{EFEFEF}91.54                                      \\ \cline{2-7} 
\multicolumn{1}{|c|}{\multirow{-3}{*}{Hebrew - HTB}}   & DEP                  & \cellcolor[HTML]{EFEFEF}84.84 & \cellcolor[HTML]{EFEFEF}64.06          & \cellcolor[HTML]{EFEFEF}71.81 & \cellcolor[HTML]{EFEFEF}\textbf{79.16} & \cellcolor[HTML]{EFEFEF}N/A                                        \\ \hline
\multicolumn{1}{|c|}{}                                 & SEG                   & 100.00                        & 94.58                                  & 97.88                         & \textbf{98.57}                         & 97.99                                                              \\ \cline{2-7} 
\multicolumn{1}{|c|}{}                                 & POS                   & \cellcolor[HTML]{EFEFEF}96.02 & \cellcolor[HTML]{EFEFEF}91.10          & \cellcolor[HTML]{EFEFEF}94.22 & \cellcolor[HTML]{EFEFEF}\textbf{94.26} & \cellcolor[HTML]{EFEFEF}93.3                                       \\ \cline{2-7} 
\multicolumn{1}{|c|}{\multirow{-3}{*}{Arabic - PADT}}  & DEP                  & \cellcolor[HTML]{EFEFEF}81.89 & \cellcolor[HTML]{EFEFEF}73.55          & \cellcolor[HTML]{EFEFEF}72.16 & \cellcolor[HTML]{EFEFEF}\textbf{78.72} & \cellcolor[HTML]{EFEFEF}N/A                                        \\ \hline
\multicolumn{1}{|c|}{}                                 & SEG                   & 100.00                        & 98.31                                  & 98.07                         & \textbf{98.43}                         & 97.51                                                              \\ \cline{2-7} 
\multicolumn{1}{|c|}{}                                 & POS                   & \cellcolor[HTML]{EFEFEF}93.44 & \cellcolor[HTML]{EFEFEF}\textbf{93.98} & \cellcolor[HTML]{EFEFEF}92.76 & \cellcolor[HTML]{EFEFEF}93.39          & \cellcolor[HTML]{EFEFEF}91.89                                      \\ \cline{2-7} 
\multicolumn{1}{|c|}{\multirow{-3}{*}{Turkish - IMST}} & DEP                  & \cellcolor[HTML]{EFEFEF}66.92 & \cellcolor[HTML]{EFEFEF}64.27          & \cellcolor[HTML]{EFEFEF}63.90 & \cellcolor[HTML]{EFEFEF}\textbf{64.40} & \cellcolor[HTML]{EFEFEF}N/A                                        \\ \hline
\end{tabular}
}
\caption{POS tagging and Dependency Parsing Results. \textit{SEG}, \textit{POS}, and \textit{DEP} stands for segmentation, POS tagging and Dependency Parsing F1-scores.}
\label{pos_table}
\end{table}

\begin{table}[t]
\centering
\scalebox{0.67}{
\begin{tabular}{cc|c|c|c|c|c|c|}
\cline{3-8}
\multicolumn{1}{l}{}                         & \multicolumn{1}{l|}{} & \multicolumn{1}{l|}{Oracle}   & \multicolumn{1}{l|}{UDPipe}   & Stanza                        & CATS                                   &                                                                    &                                                                     \\ \cline{3-6}
\multicolumn{1}{l}{}                         & \multicolumn{1}{l|}{} & \multicolumn{4}{c|}{BERT Token Classification}                                                                                         & \multirow{-2}{*}{\begin{tabular}[c]{@{}c@{}}J\\ CATS\end{tabular}} & \multirow{-2}{*}{\begin{tabular}[c]{@{}c@{}}JS\\ CATS\end{tabular}} \\ \hline
\multicolumn{1}{|c|}{}                       & SEG                   & 100.00                        & 89.23                         & 92.20                         & \textbf{96.72}                         & 96.02                                                              & 95.31                                                               \\ \cline{2-8} 
\multicolumn{1}{|c|}{\multirow{-2}{*}{Test}} & NER                   & \cellcolor[HTML]{EFEFEF}80.07 & \cellcolor[HTML]{EFEFEF}70.79 & \cellcolor[HTML]{EFEFEF}67.90 & \cellcolor[HTML]{EFEFEF}\textbf{74.46} & \cellcolor[HTML]{EFEFEF}48.36                                      & \cellcolor[HTML]{EFEFEF}60.23                                       \\ \hline
\multicolumn{1}{|c|}{}                       & SEG                   & 100.00                        & 88.47                         & 92.00                         & 95.48                                  & \textbf{95.80}                                                     & 95.38                                                               \\ \cline{2-8} 
\multicolumn{1}{|c|}{\multirow{-2}{*}{Dev}}  & NER                   & \cellcolor[HTML]{EFEFEF}78.69 & \cellcolor[HTML]{EFEFEF}68.13 & \cellcolor[HTML]{EFEFEF}66.88 & \cellcolor[HTML]{EFEFEF}\textbf{71.34} & \cellcolor[HTML]{EFEFEF}45.75                                      & \cellcolor[HTML]{EFEFEF}61.57                                       \\ \hline
\end{tabular}
}
\caption{Hebrew SEG and NER F1-Scores. %
}
\label{ner_table}
\end{table}

Tables \ref{pos_table} and  \ref{ner_table} present our pipeline results on POS, Dependency parsing and NER labeling respectively, for the various segmentation possibilities. On any language that our model achieved a meaningful segmentation improvement (e.g., Hebrew, Arabic), an increase in downstream task results was also obtained. These results  confirm the claim that segmentation mistakes indeed severely contaminate the downstream tasks.

In our joint model results  in Tables \ref{pos_table} and  \ref{ner_table}, we observe that, contrary to previous studies, the joint model did not improve  the segmentation accuracy, nor the labeling score over the pipeline %
(yet the sentence embeddings in JS-CATS improved labeling results substantially compared to J-CATS).
 The  results observed here bring up again the question of `joint versus pipeline'  in the neural era, and present an opportunity to  investigate  more sophisticated joint segmentation-and-labeling models that  extend  the  proposed architecture in new ways.

A manual error analysis we performed on  100 segmented sentences is presented in Table \ref{error_analysis}. We see that UDPipe has a higher tendency towards under-segmentation both on prefixes and suffixes. Secondly, though {\em CATS BERT} resulted in far fewer errors, the model presents unnecessary artifacts (10.2\%) which are caused by the generative nature of our model. Further study on avoiding such artifacts might increase results  further.

\begin{table}[h]
\centering
\scalebox{0.7}{
\begin{tabular}{l|c|c|c|}
\cline{2-4}
                                              & UDPipe       & CATS Baseline   & CATS BERT    \\ \hline
\multicolumn{1}{|l|}{Over-seg. prefix}  & 10.78\% (11) & 38.33\% (23) & 32.62\% (16) \\
\multicolumn{1}{|l|}{Under-seg. prefix} & 75.4\% (77)  & 54.99\% (33) & 50.98\% (25) \\ \hline
\multicolumn{1}{|l|}{Over-seg. suffix}  & 1.96\% (2)   & 1.67\% (1)   & 6.12\% (3)   \\
\multicolumn{1}{|l|}{Under-seg. suffix} & 9.8\% (10)   & 3.33\% (2)   & 0\% (0)      \\ \hline
\multicolumn{1}{|l|}{Model artifacts}         & 1.9\%(2)     & 1.67\% (1)   & 10.2\% (5)   \\ \hline
\multicolumn{1}{|l|}{Total errors}            & 100\% (102)  & 100\% (60)   & 100\% (49)   \\ \hline
\end{tabular}
}
\caption{Error analysis of 100 predicted sentence analyses (number of errors in brackets). We provide further breakdown in the supplementary material.}
\label{error_analysis}
\end{table}

\section{Conclusion}

We present a simple, effective and accurate  neural segmentation model that by combining character-level sequence-to-sequence modeling with pretrained contextualized representations can effectively cope with  complex and ambiguous segmentation of multi-word tokens.
The model achieves state-of-the-art segmentation results on various languages, and leads to substantial improvements on key tasks down the pipeline. These results opens the door for (pre-)training large language models on (pre-)segmented data rather than on raw tokens, to yield even further improvements on natural language understanding on such languages.

\bibliography{anthology,custom}
\bibliographystyle{acl_natbib}

\end{document}